\documentclass[letterpaper,10pt,conference]{ieeeconf}
\IEEEoverridecommandlockouts
\usepackage{amsmath,amssymb}
\usepackage{graphicx}
\usepackage{booktabs}
\usepackage{multirow}
\usepackage{array}
\usepackage{xcolor}
\usepackage{url}
\usepackage{balance}
\usepackage{float}
\usepackage{cuted}

\preCutedStrip={\vskip -22pt}

\postCutedStrip={\vskip -8pt}

\usepackage{xspace}
\usepackage{makecell}
\usepackage{adjustbox}
\usepackage{capt-of}
\usepackage{cite}

\makeatletter
\let\NAT@parse\undefined
\makeatother

\definecolor{academicblue}{RGB}{0,70,140}

\usepackage[
colorlinks=true,
linkcolor=academicblue,
citecolor=academicblue,
urlcolor=academicblue
]{hyperref}

\makeatletter
\def\@IEEEtablecaptionsepspace{\vskip -2pt}
\makeatother

\usepackage{etoolbox}

\makeatletter
\patchcmd{\@makecaption}
{\\{\footnotesize\scshape #2}}
{\\[-3pt]{\footnotesize\scshape #2}}
{}{}
\makeatother

\makeatletter
\def\@IEEEfigurecaptionsepspace{\vskip 0pt}
\makeatother

\makeatletter
\def\@IEEEauthorblockconfadjspace{-0.65em}
\makeatother

\title{\LARGE \bf HIRE: History-Conditioned Interaction Reasoning and High-Rate Execution for Visually Aliased Precision Manipulation}

\author{%
	Rongji Li$^{1,2,3,4}$,
	Wenhao He$^{4,\ddagger}$,
	Cewu Lu$^{4,5}$,
	Xingyu Chen$^{3,\dagger}$,
	Xu-Yao Zhang$^{1,2,3,\dagger}$%
	\\[0.5em]
	\parbox{\textwidth}{\centering\small
		$^{1}$MAIS, Institute of Automation, Chinese Academy of Sciences, Beijing, China\\
		$^{2}$School of Artificial Intelligence, University of Chinese Academy of Sciences, Beijing, China\\
		$^{3}$Zhongguancun Academy, Beijing, China\\
		$^{4}$Noematrix\\
		$^{5}$Shanghai Jiao Tong University, Shanghai, China\\[0.3em]
		$^{\dagger}$Corresponding Authors
		\qquad
		$^{\ddagger}$Project Leader%
	}%
}

\begin{document}
\maketitle
\thispagestyle{empty}
\pagestyle{empty}

\begin{strip}
	\centering
	\includegraphics[width=\textwidth]{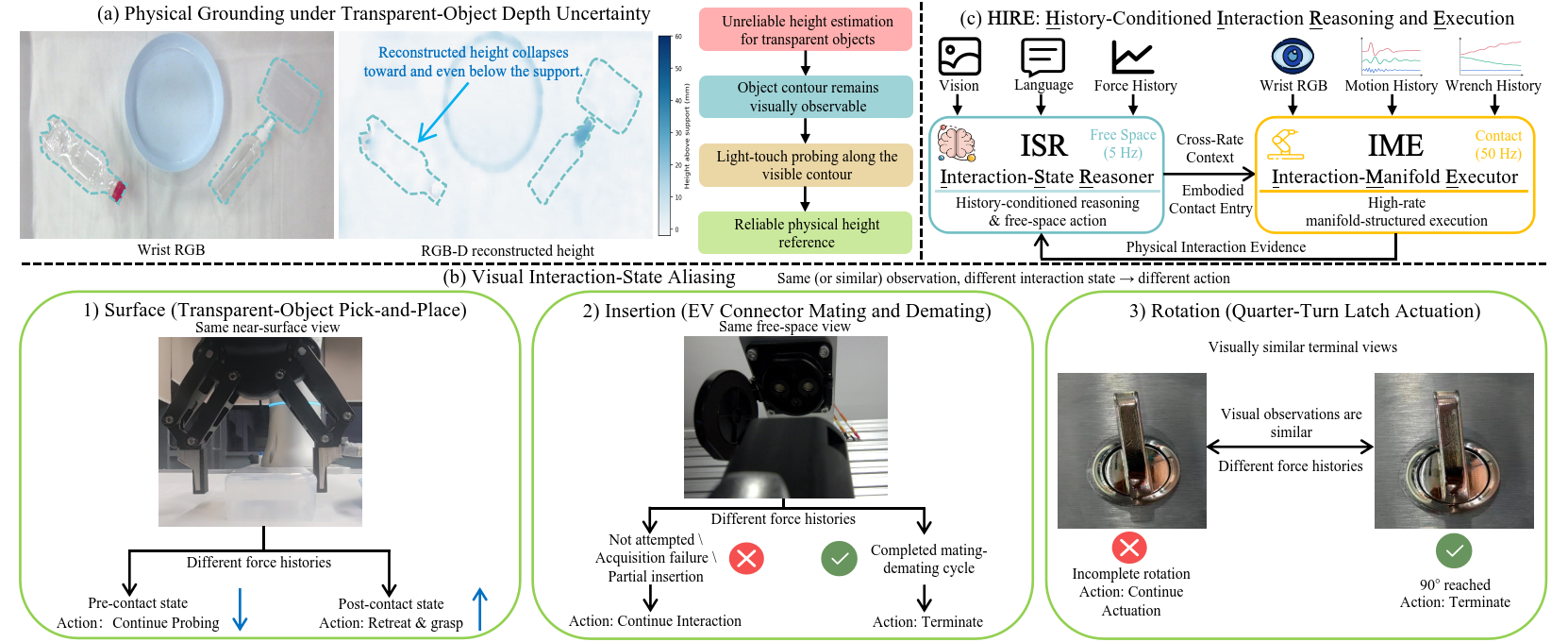}
	\captionof{figure}{
		(a) Transparent-object depth uncertainty motivates light-touch physical grounding.
		(b) Visual interaction-state aliasing: similar observations can require different actions under different physical histories.
		(c) HIRE couples history-conditioned reasoning with high-rate execution in a cross-rate loop.
	}
	\label{fig:teaser}
\end{strip}

\begin{abstract}
	
	Precision manipulation with contact-critical interactions is often history-dependent: visually similar observations can correspond to different latent interaction states and therefore require different actions, while small execution errors can alter task outcomes.
	Policies relying on the current visual observation alone cannot resolve such ambiguity; force-aware and memory-augmented methods enrich physical or temporal context, while reactive high-rate policies improve local contact response, yet long-horizon temporal reasoning and precision execution remain largely decoupled in existing methods, limiting reliable progression in visually aliased precision manipulation.
	To bridge this gap, we introduce History-Conditioned Interaction Reasoning and Execution (HIRE), a cross-rate framework comprising a history-conditioned Interaction-State Reasoner (ISR) and a high-rate Interaction-Manifold Executor (IME).
	ISR encodes ordered wrench history with a temporal wrench encoder and Force Perceiver as persistent physical evidence for state-consistent action generation, while IME structures contact-critical motion into intrinsic progress and transverse correction for precise execution; their cross-rate loop allows the resulting physical traces to inform subsequent reasoning.
	In real-robot experiments across surface, insertion, and rotational interactions, HIRE achieves at least 90\% completion across all evaluated task stages while improving interaction-state disambiguation, execution precision, and generalization.
	More broadly, HIRE provides a unified reasoning--execution perspective on precision manipulation under history-dependent partial observability, where physical interaction both realizes task intent and reveals latent-state evidence for future decisions.
	Code will be released upon publication.
	
\end{abstract}

\section{Introduction}

Recent generalist robot policies and vision-language-action (VLA) models have substantially advanced semantic generalization and visuomotor control across tasks and embodiments \cite{team2024octo,kim2024openvla,black2024pi_0,intelligence2025pi_}.
Nevertheless, precision manipulation involving critical physical interaction often violates a basic premise of observation-conditioned control: the current visual observation need not be a sufficient statistic for selecting the next action \cite{chung2026rethinking,nguyen2024leveraging}. Distinct physical interaction histories may lead to perceptually similar configurations while implying different latent interaction states and, consequently, different actions. We refer to this history-dependent partial observability as visual interaction-state aliasing.

Fig.~\ref{fig:teaser} illustrates this problem across representative contact-critical interactions.
Transparent-object pick-and-place provides a concrete example: although the object contour can remain visually discernible, transparent surfaces can corrupt RGB-D depth and reconstructed height \cite{cai2023consistent,wang2025transdiff}, making it difficult to grasp at the correct height, as shown in Fig.~\ref{fig:teaser}(a).
Rather than attempting to recover the corrupted point cloud, we use controlled light-touch probing along the visible contour to establish a reliable physical height reference.
Depending on the realized object height, contact may occur with either the support surface or the object surface; a short clearance transition unloads contact and moves the gripper toward a grasp-ready configuration.
This creates a history-dependent decision point: nearly identical near-surface observations may indicate either that no height reference has yet been established, requiring continued probing, or that contact has already occurred, requiring retreat toward grasping. 
As shown in Fig.~\ref{fig:teaser}(b), the same aliasing arises in insertion and rotational interactions, where visually similar observations may follow incomplete or completed interaction and therefore require continuation or termination.
The relevant uncertainty is therefore not geometric appearance alone, but what physical interaction has occurred.

Vision-conditioned visuomotor policies infer actions primarily from the current visual observation \cite{chi2025diffusion,zhao2023learning,brohan2022rt}, and therefore cannot resolve perceptually similar configurations whose distinction lies in preceding physical interaction.
Force-aware policies augment action generation with wrench or torque feedback \cite{yu2026forcevla,zhang2025ta,li2026forcevla2}, but instantaneous physical signals alone may no longer reveal what occurred once contact has ended.
Memory-augmented policies extend temporal context beyond the current observation \cite{shi2026memoryvla,li2026fm}, while reactive or hierarchical visual--tactile policies improve short-horizon contact responsiveness through high-rate physical feedback \cite{xue2025reactive,fang2026force}.
However, long-horizon interaction-state reasoning and precision-critical execution remain largely decoupled across these methods, limiting reliable task progression in visually aliased precision manipulation.
The key challenge is therefore to preserve physical interaction evidence across contact and free-space phases while coupling history-dependent reasoning with precise high-rate execution.

To address this challenge, we introduce \textbf{History-Conditioned Interaction Reasoning and Execution (HIRE)}, a cross-rate framework that couples force-history-conditioned reasoning with high-rate execution for visually aliased precision manipulation.
Our key insight is to treat the temporally ordered wrench trajectory not merely as instantaneous control feedback, but as persistent physical interaction evidence for state-consistent reasoning. 
As summarized in Fig.~\ref{fig:teaser}(c), the \textbf{Interaction-State Reasoner (ISR)} fuses a structured wrench-history representation with visual-language context to generate state-consistent free-space actions.
During contact-critical phases, the \textbf{Interaction-Manifold Executor (IME)} realizes the inferred interaction intent through high-rate, manifold-structured execution. 
The two modules form a cross-rate closed loop: ISR provides cross-rate interaction context to IME, while the physical trace produced through interaction with the environment becomes evidence for subsequent ISR decisions.

Our contributions are threefold:
\begin{itemize}
	\item We formulate visual interaction-state aliasing as a history-dependent partial-observability problem in precision manipulation and introduce ISR, which uses temporally ordered wrench history as persistent physical evidence for state-consistent action generation.
	
	\item We develop HIRE, a cross-rate reasoning--execution framework that couples history-aware ISR with high-rate manifold-structured IME, allowing physical interaction outcomes to inform subsequent state reasoning.	
	
	\item We validate HIRE through real-robot experiments across three complementary interaction regimes---transient surface interaction, constrained insertion, and sustained rotational interaction---demonstrating consistent improvements in stage-wise task completion, interaction-state disambiguation and contact-critical execution precision while exhibiting robust generalization.

\end{itemize}

\section{Related Work}

\subsection{Multimodal and Memory-Augmented Robot Policies}

Vision-conditioned robot policies and VLA models have achieved strong manipulation performance and generalization \cite{jiang2023vima,brohan2023rt,o2024open,belkhale2024rt,li2024cogact,team2024octo,kim2024openvla,black2024pi_0,intelligence2025pi_}.
However, when perceptually similar configurations arise from different preceding physical interactions, visual observations alone may not provide sufficient evidence for subsequent task progression.
Recent multimodal policies therefore incorporate tactile, force, or torque feedback to improve physical interaction perception and regulation \cite{he2025foar,huang20243d,zhang2025kinedex,huang2026tactile,zhang2026craft,yu2026forcevla,zhang2025ta,li2026forcevla2}.
These approaches primarily exploit physical feedback during ongoing interaction, while the task-relevant evidence from a preceding interaction may no longer be available from instantaneous or short-horizon feedback.
Memory-augmented policies extend temporal context through perceptual, action, retrieval-based, or physical histories \cite{li2025map,sridhar2026scaling,shah2026memory,li2026global,shi2026memoryvla,li2026fm,yang2026temporalflow}.
Perceptual and action memories may remain ambiguous when the relevant event is visually indistinguishable, while force-based memory preserves mechanically grounded history but still leaves precise execution challenging under rapidly evolving physical interaction.
HIRE instead couples force-history-conditioned reasoning with high-rate physical execution in a closed loop, enabling precise manipulation under visual interaction-state aliasing.

\subsection{High-Rate Interaction Execution}

Recent learning-based policies increasingly exploit tactile or force feedback for responsive physical execution and force--motion regulation \cite{liu2025forcemimic,wu2025tacdiffusion,xue2025reactive,chen2026implicitrdp,fang2026force}.
These approaches improve local responsiveness and physical regulation through reactive, hierarchical, or hybrid control, but typically leave task-relevant interaction geometry implicit in generic action or motion representations.
HIRE differs not primarily in execution frequency, but in how high-rate interaction is structured: IME organizes execution around a local interaction manifold, separating intrinsic interaction progress from transverse correction while jointly regulating reference wrench and force--position control subspaces.
This structure supports precise execution across diverse interaction geometries by coordinating interaction progress with correction of interaction-induced deviations.

\section{Problem Formulation}

\subsection{Visual Interaction-State Aliasing}

We consider contact-critical precision manipulation under a task instruction \(\ell\).
Let \(I_t\) denote the current visual observation, \(z_t\) the latent interaction state, and \(a_t^\star\) the desired action.
The latent state \(z_t\) depends on task-relevant outcomes of preceding physical interaction and may not be recoverable from the current visual observation alone.

We define visual interaction-state aliasing as the existence of two decision points under the same task instruction satisfying
\begin{equation}
	I_t \approx I_{t'}, \qquad
	z_t \neq z_{t'}, \qquad
	a_t^\star \neq a_{t'}^\star ,
	\label{eq:interaction_aliasing}
\end{equation}
Thus, visually similar observations can require different actions because they correspond to different latent interaction states shaped by preceding physical interaction.

\subsection{Contact-Critical Execution Requirements}

Contact-critical phases impose a distinct execution challenge.
Small execution errors can change whether contact is established, maintained, or mechanically completed, making closed-loop physical interaction sensitive to both control latency and execution accuracy \cite{wu2025tacdiffusion}.
These phases therefore demand a faster control timescale and tighter execution tolerances than task-level visual reasoning.

Accordingly, we study visually aliased precision manipulation with contact-critical interactions, where reliable task completion requires both resolving history-dependent interaction-state ambiguity and maintaining responsive, precise control during contact-critical execution.

\begin{figure*}[t]
	\centering
	\includegraphics[width=\textwidth]{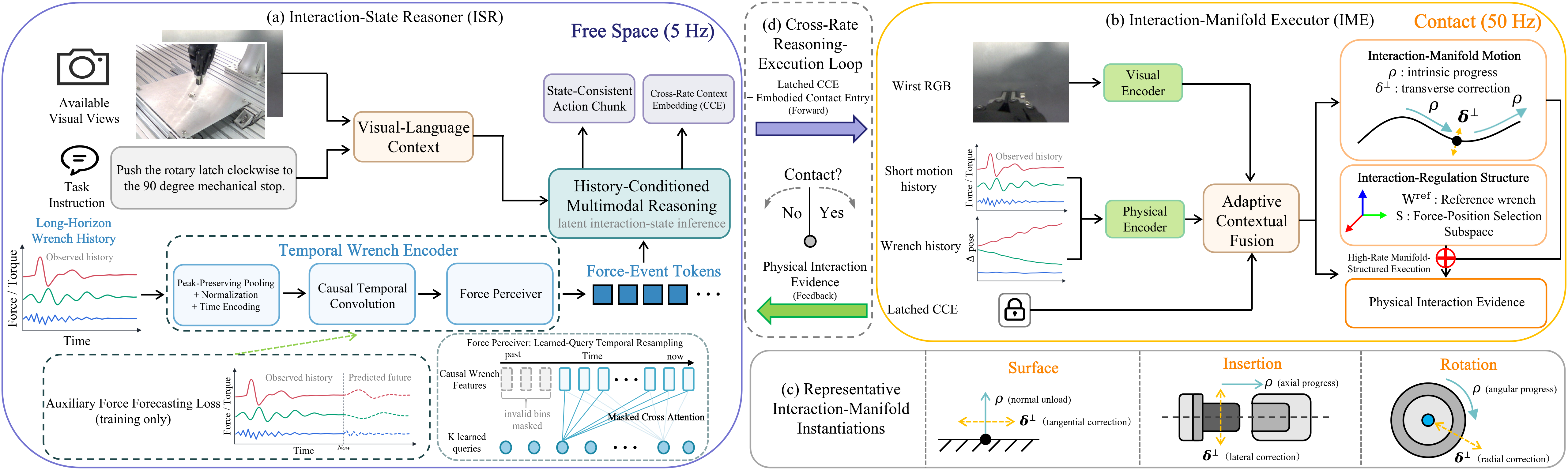}
	\caption{
		Overview of HIRE.
		(a) ISR compresses long-horizon wrench traces into force-event tokens for history-conditioned action generation.
		(b) IME combines local sensing with a latched CCE for high-rate manifold-structured interaction execution.
		(c) Representative interaction-manifold instantiations for surface, insertion, and rotational interactions.
		(d) Cross-rate coupling passes the CCE and embodied contact-entry condition forward, while measured physical interaction evidence returns to subsequent ISR reasoning.
	}
	\label{fig:method}
\end{figure*}

\section{Method}
\label{sec:method}

HIRE realizes history-dependent interaction-state reasoning and contact-critical execution through the architecture shown in Fig.~\ref{fig:method}.
At \(5\,\mathrm{Hz}\), the Interaction-State Reasoner (ISR) integrates available visual observations, language, and temporally ordered physical interaction history to generate state-consistent continuous actions together with a Cross-Rate Context Embedding (CCE).
During contact-critical intervals, the Interaction-Manifold Executor (IME) operates at \(50\,\mathrm{Hz}\) and combines local sensory feedback with the latched CCE for manifold-structured interaction execution.
The two pathways form a cross-rate closed loop: ISR determines state-consistent task progression before and after contact, while IME realizes contact-critical interaction and produces new physical evidence for subsequent ISR decisions.

Let \(\mathcal{I}_{j}^{E}=[\tau_{j}^{\mathrm{on}},\tau_{j}^{\mathrm{off}})\) denote the \(j\)-th contact-critical interval during which IME has execution authority.
The common HIRE interface is
\begin{equation}
	\begin{aligned}
		\left(\mathbf{A}_{t:t+H}^{R},\mathbf{c}_{t}^{R\rightarrow E}\right)
		&=
		\mathrm{ISR}_{\theta}\!\left(\mathcal{V}_{t},\ell,H_{t}^{F}\right),
		&& t\notin\bigcup_j\mathcal{I}_{j}^{E},\\
		\mathcal{U}_{t}^{E}
		&=
		\mathrm{IME}_{\psi}\!\left(I_{t}^{\mathrm{wrist}},H_{t}^{E},\bar{\mathbf{c}}_{j}^{R\rightarrow E}\right),
		&& t\in\mathcal{I}_{j}^{E},
	\end{aligned}
	\label{eq:hire_interface}
\end{equation}
where \(\mathcal{V}_{t}\) denotes the available visual views, \(H_{t}^{F}\) is the task-relevant long-horizon wrench history used by ISR, \(H_{t}^{E}\) is the short-horizon local motion--wrench history used by IME, \(\mathbf{c}_{t}^{R\rightarrow E}\) is the CCE produced by ISR, and \(\bar{\mathbf{c}}_{j}^{R\rightarrow E}\) denotes the CCE latched for the corresponding contact-critical interval.

\subsection{Force-History-Conditioned Interaction-State Reasoner}

As shown in Fig.~\ref{fig:method}(a), ISR treats the preceding wrench trajectory as a temporally grounded record of physical interaction rather than instantaneous feedback.
Interaction-state inference remains implicit in continuous action generation rather than being implemented as a separate phase, success, or termination classifier.
Consequently, perceptually similar observations can induce different actions when their preceding physical interaction histories provide different evidence about the latent interaction state.

At each ISR decision, we construct a causal task-relevant window from the \(100\,\mathrm{Hz}\) six-axis wrench stream.
The history is reduced to \(25\,\mathrm{Hz}\) by peak-preserving temporal pooling: within each temporal bin, we retain the complete six-axis wrench sample with the largest translational-force magnitude, preserving short interaction transients without averaging away their temporal signatures.
The resulting wrench sequence is normalized using separate physical scales for force and moment and augmented with a log-spaced sinusoidal encoding of elapsed time relative to the current decision.
A causal temporal convolution subsequently produces local physical features \(\mathbf{X}_{t}^{F}=\{\mathbf{x}_{t,i}^{F}\}_{i=1}^{L_t}\), exposing short-range wrench variations while preserving causal temporal ordering.

We then compress the variable-length temporal history into a fixed token budget using a Force Perceiver.
Starting from \(K\) learned latent queries \(\mathbf{Q}^{(0)}\in\mathbb{R}^{K\times d}\), each resampling layer retrieves evidence from the full set of valid causal wrench features.
Suppressing the attention-head index for clarity, the resampling operation is
\begin{equation}
	\begin{aligned}
		\mathbf{A}_{t}^{(l)}
		&=
		\operatorname{softmax}\!\left(
		\frac{
			\left(\mathbf{Q}^{(l)}W_Q\right)
			\left(\mathbf{X}_{t}^{F}W_K\right)^{\top}
		}{
			\sqrt{d_h}
		}
		+
		\mathbf{M}_{t}^{F}
		\right),\\
		\widetilde{\mathbf{Q}}^{(l)}
		&=
		\mathbf{Q}^{(l)}
		+
		\mathbf{A}_{t}^{(l)}
		\left(\mathbf{X}_{t}^{F}W_V\right),\\
		\mathbf{Q}^{(l+1)}
		&=
		\widetilde{\mathbf{Q}}^{(l)}
		+
		\operatorname{FFN}\!\left(\operatorname{LN}(\widetilde{\mathbf{Q}}^{(l)})\right),\\
		\mathbf{Z}_{t}^{F}
		&=
		\mathbf{Q}^{(L)}W_O .
	\end{aligned}
	\label{eq:force_perceiver}
\end{equation}
Here, \(\mathbf{A}_{t}^{(l)}\in\mathbb{R}^{K\times L_t}\) contains the attention weights from the \(K\) learned queries to the \(L_t\) temporal wrench features, and \(\mathbf{M}_{t}^{F}\) masks invalid temporal bins.
The resulting \(\mathbf{Z}_{t}^{F}=\{\mathbf{z}_{t,k}^{F}\}_{k=1}^{K}\) constitutes the force-event tokens, with \(K=8\) in our implementation.
Because \(K\) is independent of the history length \(L_t\), the Force Perceiver maps variable-length temporal horizons to a fixed token budget while allowing each latent query to selectively aggregate evidence from the complete causal interaction trace.
Temporal information remains accessible through the time-encoded wrench features, without requiring hand-crafted interaction-event labels.

The force-event tokens are inserted into the ISR prefix together with the available visual and language tokens and are jointly contextualized for continuous action generation.
Inspired by hierarchical contextual conditioning in prior robot policies \cite{chen2026fast,fang2026force,shi2025hi,zou2025asynchronous}, ISR additionally extracts a Cross-Rate Context Embedding (CCE) by masked mean pooling over the final multimodal prefix features.
This pooled representation provides IME with a compact summary of the visual, linguistic, and force-history context inferred immediately before contact, rather than an explicit phase label or prescribed local trajectory.

Because the primary flow-matching objective can be partially satisfied from the strong visual-language representation alone, the newly introduced force-history pathway may otherwise be under-utilized during training.
We therefore introduce a task-agnostic self-supervised future-wrench objective that directly regularizes the force-event representation.
Let \(\mathbf{W}_{t}^{+}=\{\mathbf{w}_{t,h}^{+}\}_{h=1}^{H_F}\) denote a short subsequent wrench segment and \(m_{t,h}^{+}\) indicate whether the corresponding future target is available.
A lightweight two-layer MLP \(G_F\) maps the concatenated force-event tokens to the predicted future wrench sequence,
\begin{equation}
	\begin{aligned}
		\widehat{\mathbf{W}}_{t}^{+}
		&=
		G_F\!\left(
		[\mathbf{z}_{t,1}^{F};\ldots;\mathbf{z}_{t,K}^{F}]
		\right),\\
		\mathcal{L}_{\mathrm{forecast}}
		&=
		\frac{
			\sum_{h}m_{t,h}^{+}
			\left\|
			\widehat{\mathbf{w}}_{t,h}^{+}-\mathbf{w}_{t,h}^{+}
			\right\|_{2}^{2}
		}{
			6\sum_{h}m_{t,h}^{+}+\epsilon
		},\\
		\mathcal{L}_{\mathrm{ISR}}
		&=
		\mathcal{L}_{\mathrm{flow}}
		+
		\lambda_F\mathcal{L}_{\mathrm{forecast}} .
	\end{aligned}
	\label{eq:force_forecast}
\end{equation}
Here, \(\lambda_F\) weights the auxiliary forecasting objective.
Forecasting subsequent wrench evolution provides a direct physical learning signal for the history pathway and encourages the force-event tokens to retain interaction dynamics that are informative beyond the current observation.
The forecasting branch is used only during training and requires no future wrench information at inference.

\begin{figure*}[t]
	\centering
	\includegraphics[width=\textwidth]{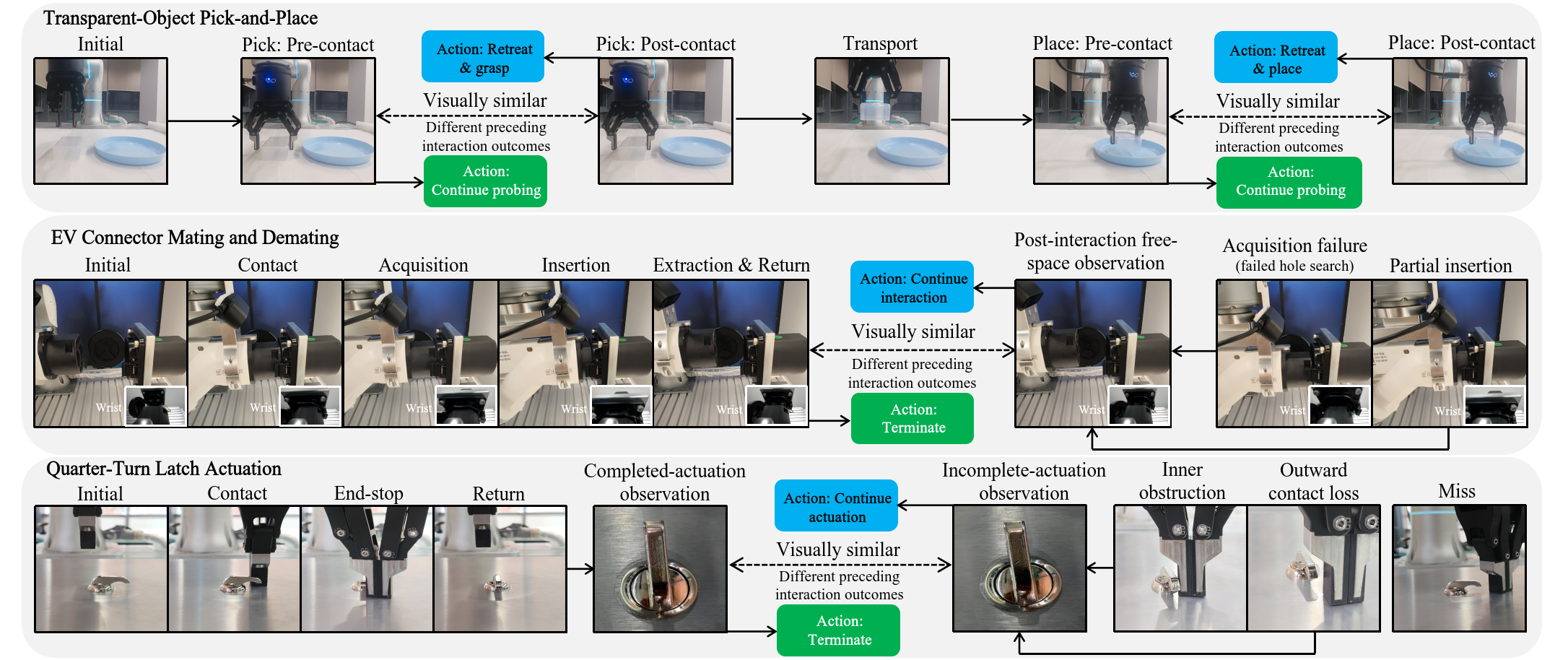}
	\caption{
		Representative real-robot execution flows and visually aliased decision states across the three evaluation regimes.
		The large external-view images are captured from closer visualization viewpoints than the policy's global-camera observations and are used only for visualization.
	}
	\label{fig:tasks}
\end{figure*}

\subsection{High-Rate Interaction-Manifold Executor}

During each contact-critical interval, IME operates at \(50\,\mathrm{Hz}\) using wrist-centric visual feedback, short-horizon relative motion--wrench history, and the latched CCE, as illustrated in Fig.~\ref{fig:method}(b).
The central design of IME is to organize high-rate execution according to the local geometry of physical interaction.

At the interaction level, we represent the structured execution command as
\begin{equation}
	\mathcal{U}_{t}^{E}
	=
	\left(
	\rho_{t},
	\boldsymbol{\delta}_{t}^{\perp},
	W_{t}^{\mathrm{ref}},
	S_{t}
	\right),
	\label{eq:ime_representation}
\end{equation}
where the intrinsic coordinate \(\rho_t\) describes progress along the local interaction manifold, while the transverse component \(\boldsymbol{\delta}_t^\perp\) captures corrective motion in directions transverse to the manifold.
Together, \((\rho_t,\boldsymbol{\delta}_t^\perp)\) form the interaction-manifold motion representation, explicitly factorizing intrinsic progress from transverse correction rather than entangling both in a single Cartesian motion prediction.
The same decomposition can therefore be instantiated across distinct interaction geometries.

Following Force Policy \cite{fang2026force}, the reference wrench \(W_t^{\mathrm{ref}}\in\mathbb{R}^{6}\) specifies the desired translational forces and rotational moments along directions selected for wrench regulation.
The selection subspace \(S_t\in\{0,1\}^{6}\) determines the control mode of each translational and rotational degree of freedom: \(S_{t,i}=1\) selects wrench regulation toward \(W_{t,i}^{\mathrm{ref}}\), whereas \(S_{t,i}=0\) leaves that direction governed by the manifold-derived motion command.
Thus, \((W_t^{\mathrm{ref}},S_t)\) defines the interaction-regulation structure that complements manifold-structured motion with hybrid force--position regulation.

All structured execution variables are learned jointly from synchronized high-rate physical demonstrations.
Let \(\hat{\rho}_t\), \(\hat{\boldsymbol{\delta}}_t^\perp\), \(\hat{W}_t^{\mathrm{ref}}\), and \(\hat{S}_t\) denote the IME predictions, with corresponding demonstration-derived targets denoted by superscript \({}^\star\).
We optimize
\begin{equation}
	\mathcal{L}_{\mathrm{IME}}
	=
	\lambda_{\rho}\mathcal{L}_{\rho}
	+
	\lambda_{\perp}\mathcal{L}_{\perp}
	+
	\lambda_{W}\mathcal{L}_{W}
	+
	\lambda_{S}\mathcal{L}_{S},
	\label{eq:ime_loss}
\end{equation}
where
\begin{equation}
	\begin{aligned}
		\mathcal{L}_{\rho}
		&=
		\operatorname{SmoothL1}
		\!\left(
		\hat{\rho}_{t},
		\rho_{t}^{\star}
		\right),\\
		\mathcal{L}_{\perp}
		&=
		\operatorname{SmoothL1}
		\!\left(
		\hat{\boldsymbol{\delta}}_{t}^{\perp},
		\boldsymbol{\delta}_{t}^{\perp\star}
		\right),\\
		\mathcal{L}_{W}
		&=
		\left\|
		\hat{W}_{t}^{\mathrm{ref}}
		-
		W_{t}^{\mathrm{ref}\star}
		\right\|_{2}^{2},\\
		\mathcal{L}_{S}
		&=
		\operatorname{BCE}
		\!\left(
		\hat{S}_{t},
		S_{t}^{\star}
		\right).
	\end{aligned}
	\label{eq:ime_losses}
\end{equation}
The first two terms supervise intrinsic interaction progress and transverse correction, while the latter two supervise the interaction-regulation variables.
This structured objective provides a common learning interface for high-rate execution across distinct interaction geometries.

Fig.~\ref{fig:method}(c) illustrates representative instantiations of the interaction-manifold representation.
For transient surface interaction, \(\rho_t\) describes normal unloading after contact, while \(\boldsymbol{\delta}_t^\perp\) captures tangential correction within the local contact plane.
For constrained insertion, \(\rho_t\) represents axial interaction progress, while \(\boldsymbol{\delta}_t^\perp\) compensates lateral misalignment.
For sustained rotational interaction, \(\rho_t\) represents monotonic angular progress around an object-relative local pivot, while \(\boldsymbol{\delta}_t^\perp\) provides bounded radial correction.
Across these distinct geometries, IME retains the same factorization of intrinsic progress, transverse correction, and hybrid force--position regulation.

\subsection{Cross-Rate Reasoning--Execution Loop}

Fig.~\ref{fig:method}(d) summarizes the bidirectional coupling between interaction-state reasoning and contact-critical execution.
The forward coupling comprises two complementary pathways.
At contact onset, the final pre-contact CCE is latched and held fixed throughout the corresponding IME interval.
In parallel, execution of the ISR action chunk physically establishes the embodied contact-entry condition from which IME begins local interaction.
The embodied contact-entry condition is neither encoded in the CCE nor transmitted as an additional symbolic message; instead, IME directly observes the realized physical configuration through its wrist-centric visual and motion--wrench feedback.

Contact evidence with hysteresis transfers execution authority from ISR to IME at contact onset and back to ISR after confirmed contact release.
The reverse pathway is mediated by physical interaction rather than an additional latent message from IME: contact-critical execution produces a measured wrench trace that is accumulated into the subsequent history \(H_{t'}^{F}\) when ISR resumes reasoning in free space.
Each contact-critical interval therefore serves a dual role, both realizing the current interaction intent and acquiring mechanically grounded evidence for future decisions.
This closes the reasoning--execution loop by allowing later visually aliased states to be resolved according to what physically occurred during the preceding interaction.

\section{Experiments}

We organize the evaluation around five questions:
\hypertarget{question:q1}{}\hyperlink{result:q1}{(Q1)}
Can HIRE reliably handle visually aliased precision manipulation across diverse interaction regimes?
\hypertarget{question:q2}{}\hyperlink{result:q2}{(Q2)}
Can ISR leverage force history for reliable interaction-state reasoning under visual aliasing?
\hypertarget{question:q3}{}\hyperlink{result:q3}{(Q3)}
Does manifold-structured IME improve contact-critical execution precision?
\hypertarget{question:q4}{}\hyperlink{result:q4}{(Q4)}
Can HIRE generalize beyond the training distribution?
\hypertarget{question:q5}{}\hyperlink{result:q5}{(Q5)}
What interaction-state evidence is retained in force history?

\subsection{Experimental Setup}

\textbf{Platform.}
We use a Flexiv Rizon 4S equipped with a Flexiv-GN01 two-finger gripper, a six-axis force/torque sensor, and two Orbbec Gemini 2 RGB-D cameras providing global and wrist views.

\textbf{Tasks.}
As illustrated in Fig.~\ref{fig:tasks}, our evaluation covers three representative regimes of visually aliased precision manipulation, instantiated as transparent-object pick-and-place for transient surface interaction, EV connector mating and demating for constrained insertion, and quarter-turn latch actuation for sustained rotational interaction.
For transparent-object pick-and-place, an analogous light-touch probing procedure is also used before placement to establish a reliable height reference, avoiding overly high release or excessive downward contact.
Surface and rotational interactions use both global and wrist observations, whereas insertion uses the wrist view only.

\textbf{Baselines.}
We compare HIRE against seven baselines: native \(\pi_{0.5}\) \cite{intelligence2025pi_}, \(\pi_{0.5}\)+naive force (current 6-D wrench concatenated with the state input), TA-VLA \cite{zhang2025ta}, ForceVLA \cite{yu2026forcevla}, FM-VLA \cite{li2026fm}, RDP \cite{xue2025reactive}, and Force Policy \cite{fang2026force}.

\begin{table*}[t]
	\centering
	\caption{Stage-wise task completion rates across three visually aliased precision-manipulation tasks.}
	\label{tab:stagewise_success}
	\small
	\setlength{\tabcolsep}{2.6pt}
	\renewcommand{\arraystretch}{1.08}
	
	\begin{adjustbox}{max width=\textwidth}
		\begin{tabular}{@{}lccccccccc@{}}
			\toprule
			
			\multicolumn{1}{c}{\multirow{2}{*}{\raisebox{-2.5pt}{\hspace{-5pt}Method}}}
			& \multicolumn{2}{c}{Transparent-Object Pick-and-Place}
			& \multicolumn{4}{c}{EV Connector Mating and Demating}
			& \multicolumn{3}{c}{Quarter-Turn Latch Actuation}
			\\
			
			\cmidrule(lr){2-3}
			\cmidrule(lr){4-7}
			\cmidrule(lr){8-10}
			
			&
			Pick
			& Place
			& Contact
			& Acquisition
			& Insertion
			& Extraction \& Return
			& Contact
			& End-stop
			& Return
			\\
			\midrule
			
			Native \(\pi_{0.5}\)~\cite{intelligence2025pi_}
			& 8/30 (26.7\%)
			& 6/30 (20.0\%)
			& 25/30 (83.3\%)
			& 13/30 (43.3\%)
			& 0/30 (0.0\%)
			& 0/30 (0.0\%)
			& 27/30 (90.0\%)
			& 7/30 (23.3\%)
			& 5/30 (16.7\%)
			\\
			
			\(\pi_{0.5}\)+Naive Force
			& 8/30 (26.7\%)
			& 7/30 (23.3\%)
			& 26/30 (86.7\%)
			& 18/30 (60.0\%)
			& 0/30 (0.0\%)
			& 0/30 (0.0\%)
			& \textbf{30/30 (100.0\%)}
			& 6/30 (20.0\%)
			& 4/30 (13.3\%)
			\\
			
			TA-VLA~\cite{zhang2025ta}
			& 13/30 (43.3\%)
			& 12/30 (40.0\%)
			& 28/30 (93.3\%)
			& 22/30 (73.3\%)
			& 0/30 (0.0\%)
			& 0/30 (0.0\%)
			& \underline{29/30 (96.7\%)}
			& 8/30 (26.7\%)
			& 6/30 (20.0\%)
			\\
			
			ForceVLA~\cite{yu2026forcevla}
			& 6/30 (20.0\%)
			& 6/30 (20.0\%)
			& 24/30 (80.0\%)
			& 17/30 (56.7\%)
			& 1/30 (3.3\%)
			& 0/30 (0.0\%)
			& \textbf{30/30 (100.0\%)}
			& 10/30 (33.3\%)
			& 10/30 (33.3\%)
			\\
			
			FM-VLA~\cite{li2026fm}
			& 15/30 (50.0\%)
			& 14/30 (46.7\%)
			& 28/30 (93.3\%)
			& 25/30 (83.3\%)
			& 6/30 (20.0\%)
			& 6/30 (20.0\%)
			& \textbf{30/30 (100.0\%)}
			& 5/30 (16.7\%)
			& 5/30 (16.7\%)
			\\
			
			RDP~\cite{xue2025reactive}
			& \underline{20/30 (66.7\%)}
			& \underline{20/30 (66.7\%)}
			& \underline{29/30 (96.7\%)}
			& \underline{26/30 (86.7\%)}
			& 8/30 (26.7\%)
			& 7/30 (23.3\%)
			& \underline{29/30 (96.7\%)}
			& \underline{13/30 (43.3\%)}
			& \underline{11/30 (36.7\%)}
			\\
			
			Force Policy~\cite{fang2026force}
			& 17/30 (56.7\%)
			& 17/30 (56.7\%)
			& 24/30 (80.0\%)
			& 21/30 (70.0\%)
			& \underline{13/30 (43.3\%)}
			& \underline{11/30 (36.7\%)}
			& \textbf{30/30 (100.0\%)}
			& 10/30 (33.3\%)
			& 9/30 (30.0\%)
			\\
			
			\midrule
			
			\textbf{HIRE (Ours)}
			& \textbf{29/30 (96.7\%)}
			& \textbf{28/30 (93.3\%)}
			& \textbf{30/30 (100.0\%)}
			& \textbf{29/30 (96.7\%)}
			& \textbf{27/30 (90.0\%)}
			& \textbf{27/30 (90.0\%)}
			& \textbf{30/30 (100.0\%)}
			& \textbf{27/30 (90.0\%)}
			& \textbf{27/30 (90.0\%)}
			\\
			
			\bottomrule
		\end{tabular}
	\end{adjustbox}
\end{table*}

\textbf{Implementation Details.}
ISR is initialized from the pretrained \(\pi_{0.5}\) backbone \cite{intelligence2025pi_}.
IME instantiates its visual and physical branches with ResNet-18 and GRU encoders, respectively, and decodes their CCE-conditioned fusion into the structured execution variables.
We further develop an automated real-robot collection-and-processing pipeline that synchronizes multimodal trajectories, performs rollout-level quality screening, and converts diverse physical interaction outcomes into training episodes, enabling rapid and repeatable dataset construction with minimal manual intervention.
The collected training dataset contains 300 real-robot demonstrations per task with synchronized visual observations, wrench histories, and executed actions.
All learning-based methods are trained on the same task demonstrations and evaluated under the same protocol.

\subsection{Results and Analysis}

\hypertarget{result:q1}{}
\textbf{HIRE sustains reliable task progression across visually aliased precision-manipulation tasks 
\hyperlink{question:q1}{(Q1)}.}
Table~\ref{tab:stagewise_success} reports cumulative stage-wise completion over 30 real-robot rollouts per task, with all methods evaluated under the same maximum execution horizon.
Across the three interaction regimes, reliable task progression requires both precise contact-critical execution and interaction-state reasoning when visually similar observations require different subsequent actions.

The stage-wise results first expose a common interaction-state reasoning failure.
In transparent-object manipulation, native \(\pi_{0.5}\), \(\pi_{0.5}\)+Naive Force, ForceVLA, and Force Policy frequently become indecisive around visually aliased near-surface configurations, where similar observations can require either continued probing or retreat toward grasping or placement.
The same ambiguity reappears near the connector interface, where pre-interaction, failed acquisition, partial insertion, and a completed mating--demating cycle can produce visually similar configurations while requiring different continuation or termination decisions.
Likewise, visually similar latch observations can correspond either to incomplete rotation or to a reached mechanical end-stop.
Without persistent evidence of the preceding physical interaction, these methods may repeatedly re-approach or withdraw around the connector, terminate before physical completion, or continue acting around an already completed latch state, consuming the execution horizon and degrading later-stage completion.

A complementary limitation arises during contact-critical execution itself.
In transparent manipulation, corrupted depth estimates can cause premature grasping above the object or excessive probing into the support surface, potentially resulting in contact overload, while low-frequency force-aware policies such as TA-VLA and FM-VLA lack a dedicated high-rate execution loop for regulating the brief contact and unloading transition.
The same bandwidth limitation becomes more severe in connector acquisition, insertion, and extraction, and during sustained latch rotation, where rapidly evolving lateral or radial deviations must be corrected continuously.
RDP alleviates this limitation through a fast tactile/force-conditioned decoder, but its high-rate actions remain conditioned on the latent action chunk generated by the slow policy; thus, fast feedback can correct local deviations within the sampled behavior, while larger changes in interaction progression remain constrained by the slow latent representation.
Force Policy likewise provides high-rate local force regulation, but its local motion is represented as a predicted motion chunk conditioned on short-horizon physical feedback.
In our constrained insertion and sustained rotation tasks, we observe that such local execution can accumulate lateral or radial deviation over prolonged contact, leading to incomplete insertion, inner obstruction, or outward contact loss.

HIRE addresses these coupled failure modes at complementary timescales while retaining a closed reasoning--execution loop.
IME maintains high-rate, manifold-structured interaction progress and transverse correction during contact-critical execution, reducing contact overload, lateral or radial drift, and contact loss, while ISR retains temporally ordered wrench history as decision-relevant physical evidence for state-consistent task progression under visually aliased observations.

\begin{figure}[H]
	\centering
	\includegraphics[width=\columnwidth]{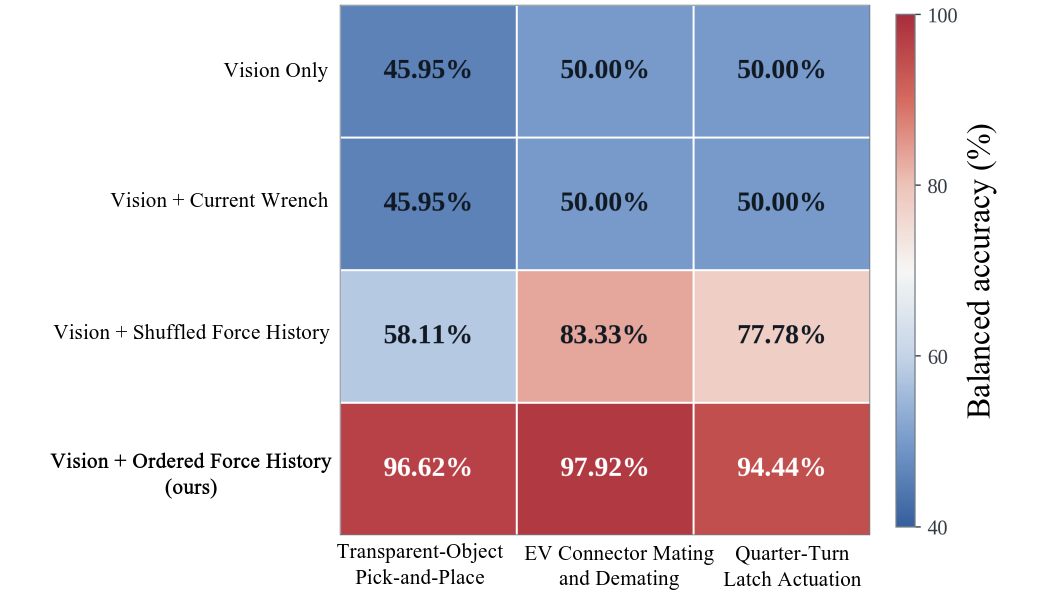}
	\caption{Balanced reasoning accuracy under different inputs.}
	\label{fig:aliasing}
\end{figure}

\hypertarget{result:q2}{}
\textbf{ISR effectively resolves visual interaction-state aliasing by integrating temporally ordered force history 
\hyperlink{question:q2}{(Q2)}.}
We evaluate ISR on visually aliased decision states using four input variants: vision only, vision with current wrench, vision with temporally shuffled force history, and vision with ordered force history. 
Balanced accuracy is computed over task-progression decisions decoded directly from the predicted continuous actions, without an auxiliary classifier.
As shown in Fig.~\ref{fig:aliasing}, vision only and current-wrench augmentation perform similarly, indicating that instantaneous observations are insufficient for reliable interaction-state reasoning.
In contrast, integrating force history through the temporal wrench encoder and Force Perceiver substantially improves accuracy across all three tasks, showing that the proposed history pathway captures decision-relevant physical interaction evidence that is unavailable from the current observation alone.
Shuffling the same history markedly degrades performance, further showing that ISR exploits not only accumulated physical evidence but also its temporal ordering.

\begin{table}[H]
	\centering
	\caption{Contact-critical execution precision during latch rotation.}
	\label{tab:ime_precision}
	\footnotesize
	\setlength{\tabcolsep}{2.0pt}
	\renewcommand{\arraystretch}{1.08}
	
	\begin{adjustbox}{max width=\columnwidth}
		\begin{tabular}{@{}lccccc@{}}
			\toprule
			
			\multirow{2}{*}{\raisebox{-10.5pt}{Method}}
			& \multicolumn{2}{c}{Motion Accuracy}
			& \multicolumn{2}{c}{Force Regulation}
			& \multirow{2}{*}{\raisebox{-10.5pt}{\makecell[c]{Contact\\Retention $\uparrow$}}} \\

			\cmidrule(lr){2-3}
			\cmidrule(lr){4-5}
			
			& \makecell[c]{Angular RMSE $\downarrow$\\($^\circ$)}
			& \makecell[c]{Radial RMSE $\downarrow$\\(mm)}
			& \makecell[c]{Normal-force RMSE $\downarrow$\\(N)}
			& \makecell[c]{Radial-force P95 $\downarrow$\\(N)}
			& \\
			
			\midrule
			
			Native \(\pi_{0.5}\)
			& 26.65
			& 2.38
			& 8.60
			& 4.46
			& 4.91\% \\
			
			\(\pi_{0.5}\)+Naive Force
			& 31.90
			& 3.23
			& 16.34
			& 7.17
			& 8.76\% \\
			
			TA-VLA
			& 17.31
			& 2.14
			& 21.11
			& 4.33
			& 5.60\% \\
			
			ForceVLA
			& 16.33
			& \underline{1.63}
			& 13.64
			& 2.49
			& 6.85\% \\
			
			FM-VLA
			& 33.88
			& 4.43
			& 23.27
			& 14.94
			& 9.08\% \\
			
			RDP
			& 12.88
			& 1.67
			& 7.02
			& \underline{2.31}
			& 10.62\% \\
			
			Force Policy
			& \underline{12.64}
			& 1.75
			& \underline{3.21}
			& 3.07
			& \underline{95.22\%} \\
			
			\midrule
			
			\textbf{HIRE (Ours)}
			& \textbf{10.05}
			& \textbf{1.20}
			& \textbf{2.83}
			& \textbf{1.00}
			& \textbf{96.59\%} \\
			
			\bottomrule
		\end{tabular}
	\end{adjustbox}
\end{table}

\hypertarget{result:q3}{}
\textbf{Manifold-structured IME improves contact-critical execution precision and stability 
\hyperlink{question:q3}{(Q3)}.}
We evaluate the contact-controlled latch-rotation phase using five complementary metrics: angular RMSE measures rotational-progress error, radial RMSE measures inward/outward drift relative to the latch pivot, normal-force RMSE measures tracking error from the \(3\,\mathrm{N}\) normal-force reference against the aluminum plate, radial-force P95 captures large radial contact loads, and contact retention measures the fraction of the rotation interval maintaining valid contact with the plate surface.
As shown in Table~\ref{tab:ime_precision}, HIRE achieves the best performance across all five metrics, demonstrating more accurate rotational progress, tighter radial correction, and stable force-regulated contact.
Force Policy attains comparable normal-force tracking and contact retention, but exhibits larger angular and radial motion errors together with higher radial-force peaks, suggesting that generic local motion prediction provides weaker coordination between interaction progress and radial correction during sustained rotation.
These results support the IME design of explicitly structuring execution into intrinsic progress and transverse correction while jointly regulating contact forces.

\begin{table}[t]
	\centering
	\caption{Generalization to unseen transparent objects.}
	\label{tab:unseen_objects}
	\footnotesize
	\setlength{\tabcolsep}{2.7pt}
	\renewcommand{\arraystretch}{1.08}
	
	\begin{adjustbox}{max width=\columnwidth}
		\begin{tabular}{@{}lccccc@{}}
			\toprule
			
			\multirow{2}{*}{\raisebox{-0.25in}{Method}}
			& \multicolumn{4}{c}{Unseen Objects}
			& \multirow{2}{*}{\raisebox{-0.25in}{Overall}} \\
			
			\cmidrule(lr){2-5}
			
			&
			\raisebox{-0.5\height}{\includegraphics[height=0.43in,keepaspectratio]{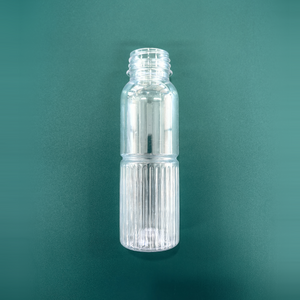}}
			&
			\raisebox{-0.5\height}{\includegraphics[height=0.43in,keepaspectratio]{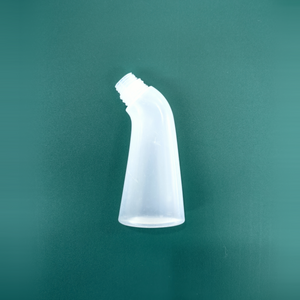}}
			&
			\raisebox{-0.5\height}{\includegraphics[height=0.43in,keepaspectratio]{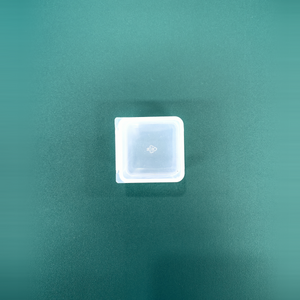}}
			&
			\raisebox{-0.5\height}{\includegraphics[height=0.43in,keepaspectratio]{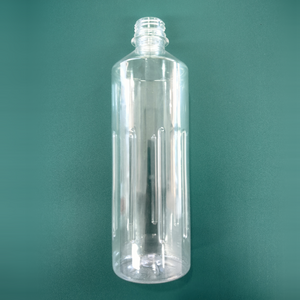}}
			&
			\\
			
			\midrule
			
			Native \(\pi_{0.5}\)
			& 1/5
			& 2/5
			& 1/5
			& 1/5
			& 5/20 \\
			
			\(\pi_{0.5}\)+Naive Force
			& 0/5
			& 1/5
			& 1/5
			& 2/5
			& 4/20 \\
			
			TA-VLA
			& 0/5
			& \underline{3/5}
			& \underline{2/5}
			& 2/5
			& 7/20 \\
			
			ForceVLA
			& 1/5
			& 2/5
			& 0/5
			& 2/5
			& 5/20 \\
			
			FM-VLA
			& 1/5
			& \underline{3/5}
			& \underline{2/5}
			& \underline{4/5}
			& 10/20 \\
			
			RDP
			& \underline{4/5}
			& \underline{3/5}
			& \underline{2/5}
			& \textbf{5/5}
			& \underline{14/20} \\
			
			Force Policy
			& 3/5
			& 2/5
			& \underline{2/5}
			& \underline{4/5}
			& 11/20 \\
			
			\midrule
			
			\textbf{HIRE (Ours)}
			& \textbf{5/5}
			& \textbf{5/5}
			& \textbf{3/5}
			& \textbf{5/5}
			& \textbf{18/20} \\
			
			\bottomrule
		\end{tabular}
	\end{adjustbox}
\end{table}

\hypertarget{result:q4}{}
\textbf{HIRE generalizes beyond the training distribution through transferable interaction reasoning and execution 
\hyperlink{question:q4}{(Q4)}.}
Table~\ref{tab:unseen_objects} reports generalization performance on transparent-object pick-and-place using unseen objects with variations in shape, height, placement, and orientation.
The pretrained \(\pi_{0.5}\) prior supports visual-semantic transfer; after light-touch establishes a physical height reference, IME adapts contact unloading to different realized surface heights, while ISR uses the resulting force history to determine whether contact has already occurred and accordingly select between continued probing and retreat toward grasping.
Together, these results show that HIRE preserves a consistent physically grounded reasoning--execution structure across unseen configurations without relying on inaccurate point-cloud height estimates.

\begin{figure}[H]
	\centering
	\includegraphics[width=\columnwidth]{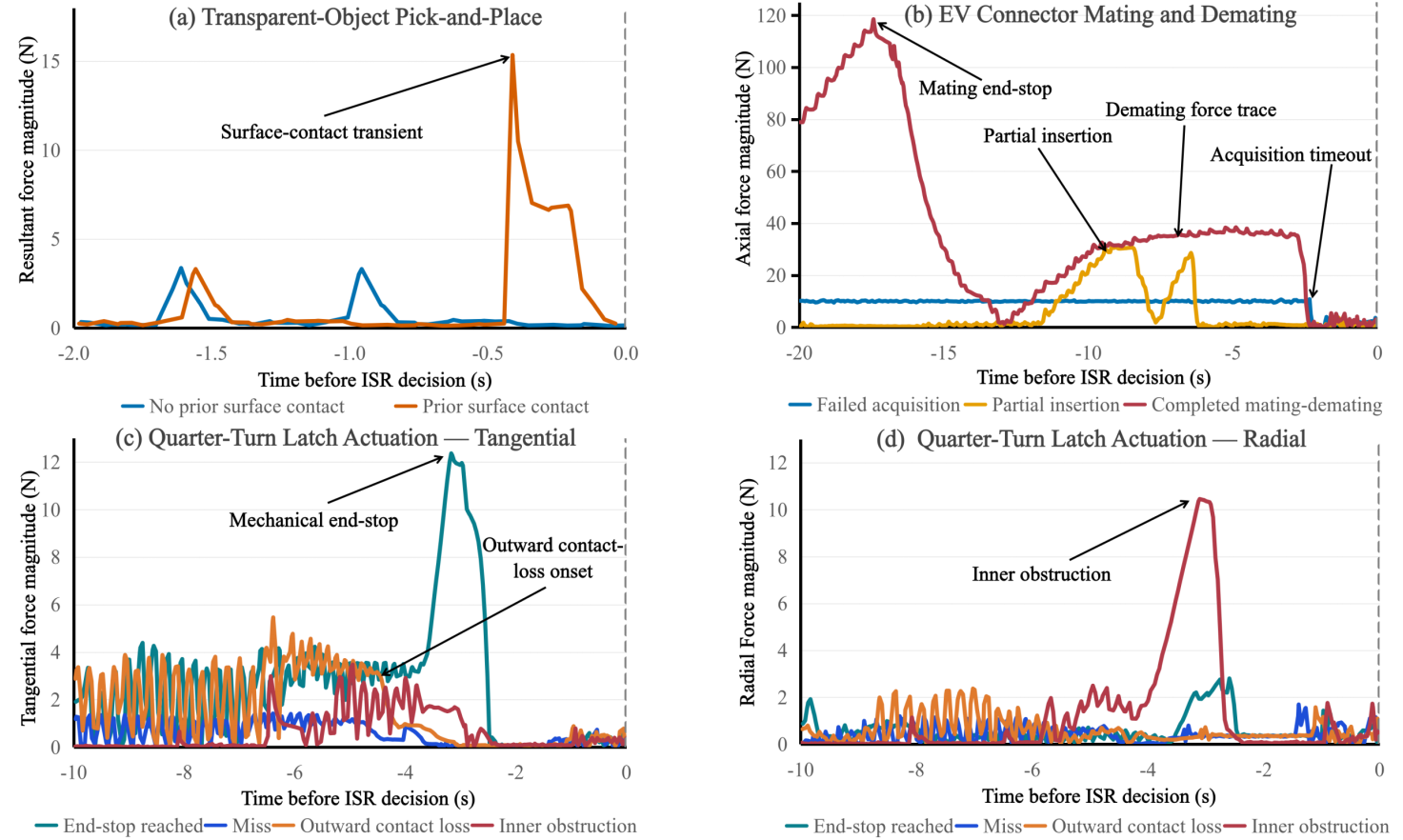}
	\caption{Wrench-history signatures under visual interaction-state aliasing.}
	\label{fig:force_history}
\end{figure}

\hypertarget{result:q5}{}
\textbf{Force history preserves mechanically grounded interaction-state evidence 
\hyperlink{question:q5}{(Q5)}.}
As shown in Fig.~\ref{fig:force_history}, visually aliased decision states across all three tasks are preceded by distinct wrench histories reflecting different realized physical interactions.
In latch actuation (Fig.~\ref{fig:force_history}(c--d)), both the mechanical end-stop and inner obstruction produce large force peaks, but the former is tangential-dominant with weak radial loading, whereas the latter is radial-dominant with weaker tangential response; outward contact loss instead produces a rapid tangential-force decay.
Our force-history pathway therefore exploits not only force magnitude, but also temporal evolution and cross-axis wrench relationships, providing ISR with physically grounded cues for resolving visually aliased states.

\section{Conclusion}

We introduce HIRE, a cross-rate framework that unifies history-conditioned interaction reasoning with high-rate manifold-structured execution for visually aliased precision manipulation.
ISR uses a temporal wrench encoder and Force Perceiver to preserve temporally ordered wrench history as physical interaction evidence for state-consistent action generation, while IME structures contact-critical execution around intrinsic interaction progress and transverse correction; their cross-rate coupling closes the loop as the resulting physical trace becomes evidence for subsequent reasoning.
Experiments across transient surface interaction, constrained insertion, and sustained rotational interaction demonstrate consistent improvements in task completion, interaction-state disambiguation, and contact-critical execution precision, together with robust generalization to unseen configurations.
More broadly, HIRE bridges task-level reasoning and contact-level physical execution by making interaction both a means of acting on the world and a source of persistent state evidence, providing a unified approach to manipulation under history-dependent partial observability.

\bibliographystyle{IEEEtran}
\bibliography{custom}

\end{document}